\documentclass[runningheads]{llncs}
\usepackage[T1]{fontenc}
\usepackage{graphicx}
\usepackage{booktabs}
\usepackage[misc]{ifsym}
\newcommand{\corr}{(\Letter)}
\usepackage{multirow}
\usepackage{amsmath} 
\usepackage{soul}
\begin{document}

\title{STTM: A New Approach Based Spatial-Temporal Transformer And Memory Network For Real-time Pressure Signal In On-demand Food Delivery}

\titlerunning{STTM For Real-time Pressure Signal Prediction}

\author{Jiang Wang \corr \and Haibin Wei \and Xiaowei Xu \and Jiacheng Shi \and Jian Nie  \and Longzhi Du \and Taixu Jiang}



\institute{Alibaba Group, Shanghai, China\\
\email{\{qiaogen.wj, haibinwei.whb, xu.xxw, yicheng.sjc, niejian.nj, du.dlz, kele.jiang\}@alibaba-inc.com}
}

\maketitle   

\begin{abstract}
On-demand Food Delivery (OFD) services have become very common around the world. For example, on the Ele.me platform, users place more than 15 million food orders every day. Predicting the Real-time Pressure Signal (RPS) is crucial for OFD services, as it is primarily used to measure the current status of pressure on the logistics system. When RPS rises, the pressure increases, and the platform needs to quickly take measures to prevent the logistics system from being overloaded. Usually, the average delivery time for all orders within a business district is used to represent RPS. Existing research on OFD services primarily focuses on predicting the delivery time of orders, while relatively less attention has been given to the study of the RPS. Previous research directly applies general models such as DeepFM, RNN, and GNN for prediction, but fails to adequately utilize the unique temporal and spatial characteristics of OFD services, and faces issues with insufficient sensitivity during sudden severe weather conditions or peak periods. To address these problems, this paper proposes a new method based on \textbf{S}patio-\textbf{T}emporal \textbf{T}ransformer and \textbf{M}emory Network (STTM). Specifically, we use a novel Spatio-Temporal Transformer structure to learn logistics features across temporal and spatial dimensions and encode the historical information of a business district and its neighbors, thereby learning both temporal and spatial information. Additionally, a Memory Network is employed to increase sensitivity to abnormal events. Experimental results on the real-world dataset show that STTM significantly outperforms previous methods in both offline experiments and the online A/B test, demonstrating the effectiveness of this method.

\keywords{On-demand food delivery \and Neural networks \and Real-time pressure signal \and Deep learning.}
\end{abstract}

\section{Introduction}
With the rapid development of mobile internet, On-demand Food Delivery (OFD) services, such as Uber Eats, DoorDash, Meituan, and Ele.me, have greatly facilitated people's lives. OFD services are crucial as they dispatch users' orders to suitable riders and ensure that the riders can deliver the food to users on time.

OFD platforms need a Real-time Pressure Signal (RPS) to measure the current pressure status of the logistics system. For example, during peak dining periods, the volume of user orders often exceeds the delivery capacity of riders, resulting in a high RPS. Similarly, during bad weather conditions, the number of riders and their delivery capacity may decrease, also leading to a high RPS. RPS is crucial, as it can guide platforms to implement operational strategies during high RPS, such as informing users in advance about potential longer delivery times, providing subsidies to users or riders, in order to enhance user experience and increase platform revenue.

\begin{figure}[htbp]
\centering
\includegraphics[width=\textwidth]{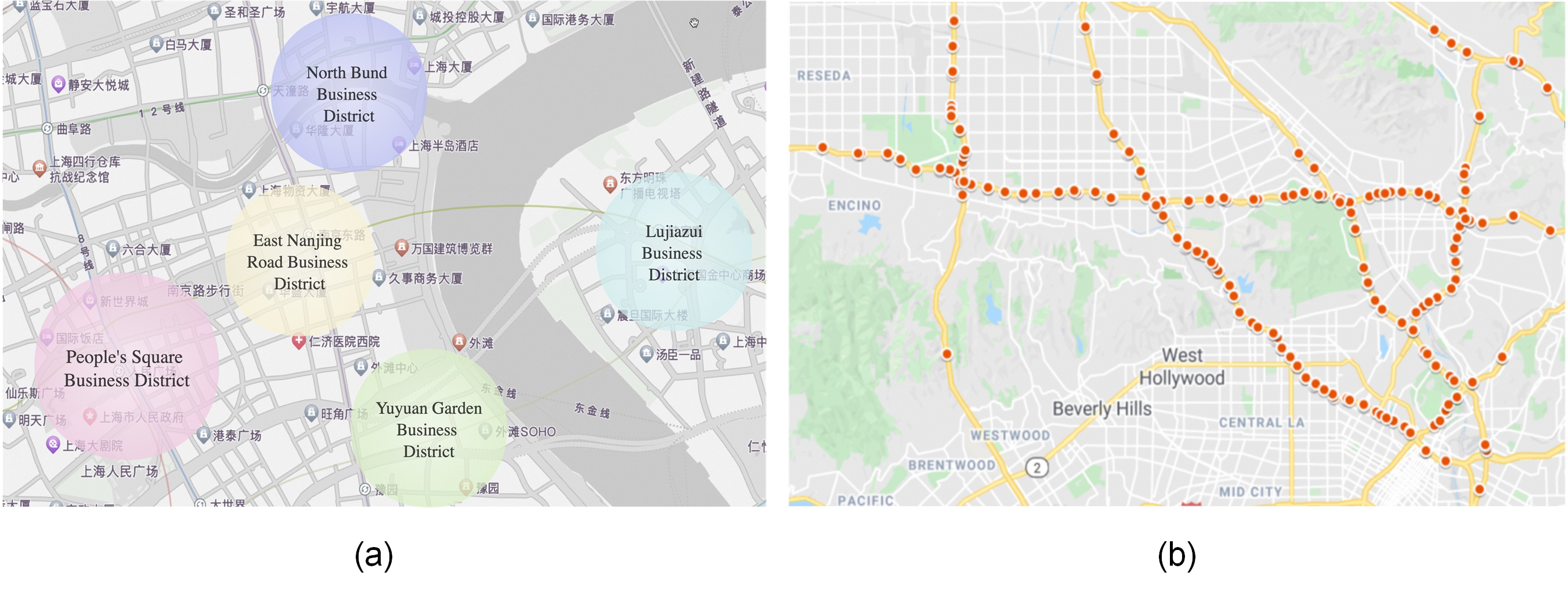}
\caption{Visualization of the spatial distribution of RPS in the OFD domain and urban traffic prediction. (a) The spatial structure of the RPS is a circular distribution of business districts, where adjacent districts will influence each other. (b) Urban traffic prediction deals with road network data sampled from sensors.} 
\label{fig1}
\end{figure}

Current research in the field of OFD mainly focuses on the prediction of food delivery time \cite{zhu2020order,gao2022applying}, while there is little research on RPS. In contrast to these studies, RPS is not based on individual orders, but rather on the business district, as the business district is the smallest unit of logistics operations. RPS can be represented by the average delivery time for all orders in a specific business district (such as People's Square or Lujiazui) in the next 5 minutes. Therefore, predicting RPS is a typical regression task. Commonly used industry methods include utilizing public domain models like the DeepFM \cite{guo2017deepfm,lian2018xdeepfm} model for direct prediction based on real-time logistics features, employing recurrent neural networks like LSTM \cite{hochreiter1997long} to encode time series features, or utilizing graph neural networks such as GAT \cite{velickovic2017graph} to encode spatial graph features. However, there are still the following challenges in directly applying such methods to the prediction of RPS:

1. RPS considers the average delivery time of all orders within a business district, and it also needs to take into account the unique spatio-temporal information of the OFD domain. RPS in various time and space exhibit high correlation, regardless of the temporal or spatial distribution. However, classic methods such as DeepFM, LSTM, and GAT can only utilize either temporal information or spatial information, but not both. Previous studies applied spatio-temporal transformers to urban traffic prediction, incorporating both temporal and spatial information. Nevertheless, due to the significant differences in spatial distributions between RPS in the OFD domain and urban traffic prediction (as shown in Fig.\@ \ref{fig1} (a).  and Fig.\@ \ref{fig1} (b)), their spatio-temporal encoding cannot be directly applied to the prediction tasks of RPS. 

2. RPS needs to quickly reflect changes when bad weather or peak periods occur. Being sensitive to these abnormal events allows the OFD platforms to quickly trigger adjustment strategies during high-pressure moments (when RPS is high), reducing logistics load and ensuring a good user experience. Previous methods do not use specific methods or structures to adapt to this requirement, so they were unable to be sensitive to sudden abnormal events.

To address these problems, this paper proposes a method based on the Spatio-Temporal Transformer and Memory Network (STTM). A new spatio-temporal Transformer structure has been designed to learn logistics characteristics in the spatio-temporal dimensions, which is very important for the RPS task. It is based on the current and historical $N$ time slices, and the logistics features of the surrounding $M$ business districts for learning and prediction. Additionally, in order to enhance the sensitivity to abnormal scenarios, a Memory Network is used to highlight the importance of features such as bad weather or peak periods. Memory Network has recently been widely applied in anomaly learning \cite{li2023traffic,wang2022event}.

The main contributions of this paper are as follows:

(1) We propose the STTM framework to systematically improve the upper limit of the ability of RPS task.

(2) We design a new Spatio-Temporal Transformer structure to learn the logistics features in the spatio-temporal dimensions and propose a relative coordinate encoder for spatial feature encoding of business distirct.

(3) We introduc Memory Network into RPS prediction to enhance sensitivity to abnormal scenarios.

(4) Experiments show that STTM has the largest improvement of 9.66\% compared to previous methods, and it has been successfully deployed on one of China's largest OFD platforms—Ele.me. Whether in offline experiments or online A/B test, STTM has shown a significant improvement.

\section{Related Work}
\subsection{Food Delivery Time Prediction}
Food delivery time prediction is the most common task in OFD services. It mainly uses machine learning methods to predict the duration of food delivery, which is the time interval from the user placing an order to the rider completing the delivery. \cite{ramesh2019doordash,jackson2019uber} first introduce a brief scheme for delivery time prediction in DoorDash and UberEats systems. Later, \cite{zhu2020order} introduce in detail the Order Fulfillment Cycle Time (OFCT) task in the Ele.me platform, and based on many logistics features under OFD domain, modeled with simple deep neural networks. Similarly, \cite{gao2022applying} also focuses on Food Preparation Time (FPT) under Meituan Waimai and propose an innovative solution. But unlike these studies, which are to predict the delivery time of each order, our RPS task is to estimate the average delivery time of all orders in each business district. Different granularities not only lead to very different changes in features, but also bring about different spatio-temporal distributions. For example, in RPS, the spatial structure we consider is more about the relationship between business districts and business districts, rather than orders and orders.

\subsection{Spatial-temporal Transformer}
The Transformer \cite{vaswani2017attention} has been widely applied in various spatio-temporal applications, such as natural language processing \cite{devlin2018bert,radford2018improving,radford2019language} and computer vision \cite{dosovitskiy2020image,caron2021emerging}. However, these tasks either involve only temporal sequences or only spatial sequences, without jointly modeling spatio-temporal information. One area with more studies that simultaneously incorporate spatio-temporal information is urban traffic prediction \cite{xu2020spatial,yan2021learning,jiang2023pdformer}, where the input comes from traffic data in the urban road network, and task-specific Transformer attention modules are designed. However, the differences between the OFD domain and urban traffic prediction field are significant, especially in the spatial structure of the input features. Urban traffic prediction deals with road network data sampled from sensors, while the spatial structure of the RPS is a circular distribution of business districts, where adjacent districts will influence each other.

\subsection{Anomaly Learning} 
Research on anomaly learning has a long history, and some recent work \cite{li2023traffic,wang2022event} uses memory networks to learn the characteristics of abnormal events. Memory Network is first introduced in the question-answering task in natural language processing \cite{chaudhari2021attentive}, where it serves as an external memory for information retrieval. Recently, it has been widely adopted in time series prediction tasks \cite{li2023traffic,tang2020joint,yao2019learning}. Abnormal events occur infrequently, so the training data is relatively limited. Therefore, \cite{yi2023deepsta} uses Memory Network to address the challenge of insufficient training data in abnormal scenarios. For the RPS task, when anomalous events such as peak periods or severe weather occur, signal needs to be quickly reflected to guide OFD platforms to quickly take corresponding measures to intervene. Therefore, this paper uses Memory Network to not only address the issue of insufficient anomalous event samples, but also to enhance the sensitivity of important features.

\section{Problem formulation}
As mentioned above, we use the average delivery time for orders placed within the next 5 minutes in the current business district to represent RPS, which is the target we want to learn:
\begin{equation}
Y=\frac{1}{N_O} \sum_{o\in O}t_{o} 
\end{equation}

where $O$ is the set of orders created in the next 5 minutes in the business district (for some large business districts, it may reach hundreds of orders), and $o$ is a certain order in $O$ , and its delivery time is represented as $t_{o}$ (that is the time interval from the user placing the order to the delivery completion by the rider). And $N_O$ is the number of orders in the $O$.

RPS prediction needs to take into account both the temporal and spatial structure of the business districts, as well as input peak periods, weather and other abnormal information. Therefore, we can divide the features into two types. The first type is features in temporal or spatio dimensions, called spatio-temporal features, denoted as $X^{A} \in {R} ^{M \times N \times D_{A} }$, where $A$ represents the first type of features, $M$ is the number of business districts in the spatial dimension, consisting of the current business district and its most adjacent $M-1$ business districts, which is a subgraph of the entire business district graph $G$ . $N$ is the number of time slices in the temporal dimension, sampled every 10 minutes. $D_A$ is the number of type $A$ features, which is composed of different types of features. The second type is mainly abnormal features, so the model must be sensitive enough to them, so we call it sensitive features, represented by $X^{B} \in {R} ^{D_{B} }$, where $B$ represents the second type of features and $D_B$ represents the number of type $B$ features. So the output can be expressed as the following formula:
\begin{equation}
\hat{Y} = \mathrm {} f_{\theta } (X^{A}, X^{B} ;G)
\end{equation}

It means that based on the spatio-temporal features $X^{A}$ of the business district graph $G$ and sensitive features $X^{B}$, learn a function $f$ with parameter $\theta$, and map the input to the predicted value $\hat{Y}$ of RPS.

\section{Method}
\subsection{The Proposed Framework} 
\begin{figure}[htbp]
\centering
\includegraphics[width=\textwidth]{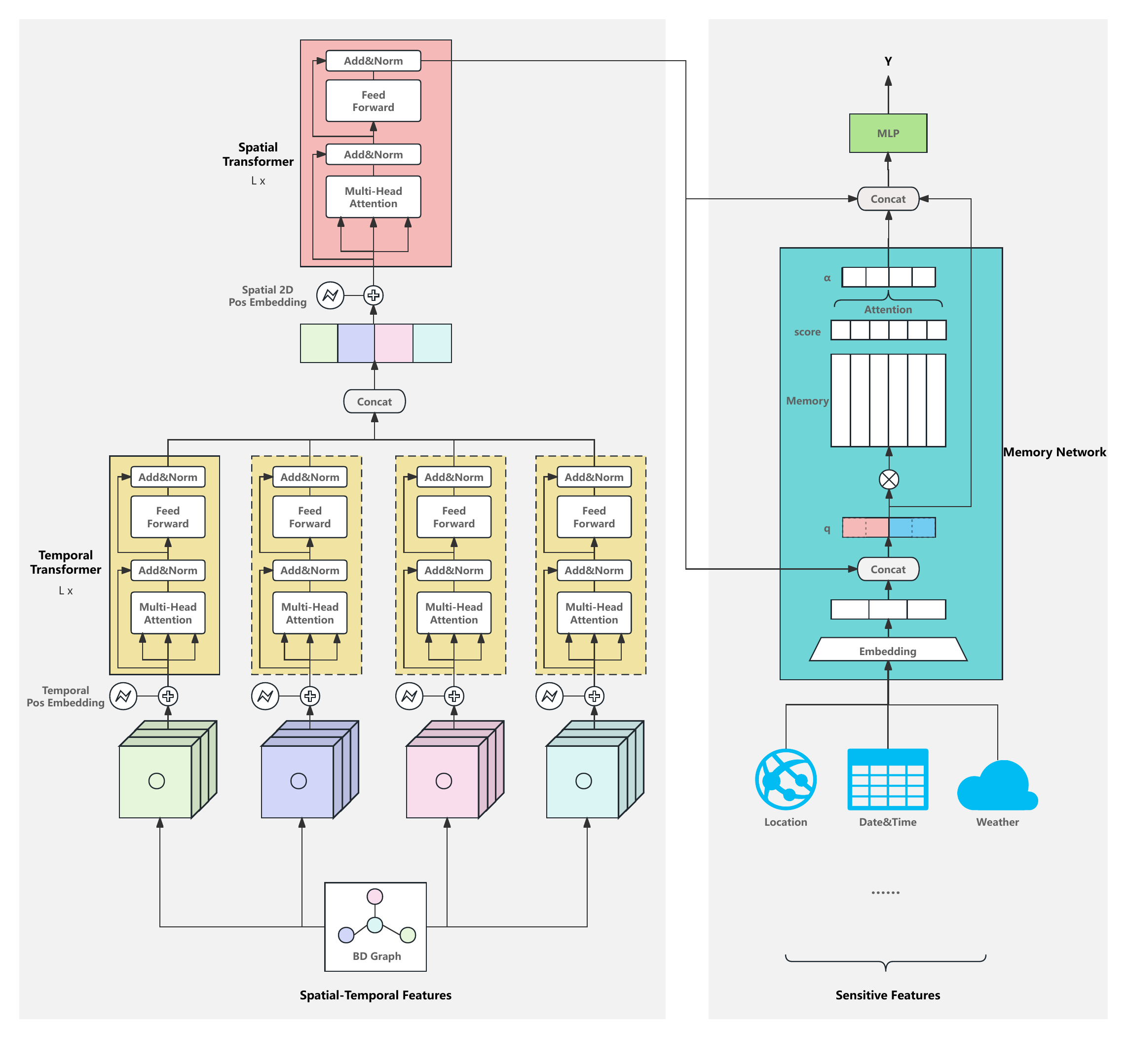}
\caption{The framework of the proposed STTM method.} 
\label{fig2}
\end{figure}

The overview architecture of STTM is shown in Fig.\@ \ref{fig2}. It mainly consists of two parts: Spatio-Temporal Transformer and Memory Network, which encode spatio-temporal features and sensitive features respectively. The Spatio-Temporal Transformer consists of two components: Temporal Transformer and Spatio Transformer. First, by processing the longitude and latitude of the business district, we calculate the nearest neighbors of the current business district, and then each business district takes its historical features as spatio-temporal features. Second, the Temporal Transformer processes the temporal features of each business district in turn, and the time information is injected through temporal position embedding. Third, the output of the Temporal Transformer for all business districts is concatenated and added with the 2D position embedding of the business district. The result is then processed by the Spatial Transformer, and its output is considered to be a summary of the spatio-temporal features. Fourth, the core of the Memory Network is the memory unit based on attention mechanism. Sensitive features such as location, time, date, weather, etc. are encoded through embedding, and then concatenated with the output of the Transformer. The resulting vector is then multiplied with the memory unit and scaled to obtain the attention distribution. Finally, we concatenate the results of the Transformer, attention distribution, and Memory Network input, and use an MLP to fit and generate the final prediction result.

\subsection{Position Embedding} 
This module is designed to inject temporal position and spatial coordinate information into the model. It mainly contains the following two components:

\subsubsection{Temporal Position Embedding}. This module is designed to add temporal information. Similar to Bert \cite{devlin2018bert}, we use absolute position encoding:
\begin{equation}
TPE_n=OneHot(n)\cdot W_{TPE}
\end{equation}
where $n\in [0, N)$ is a specific time step. $OneHot(n)\in {R} ^{1\times H}$ is the OneHot representation of $n$, with the $n$th position being 1 and all other positions being 0, $H$ is the hidden size of the Temporal Transformer. And $W_{TPE}\in {R} ^{N\times H}$ is the weight matrix of Temporal Position Embedding, randomly initialized. Therefore, $TPE_n\in {R} ^H$ and $TPE$ is eventually added to the temporal features.

\subsubsection{Spatial Position Embedding}. This module is designed to add 2D spatial coordinate information. Unlike temporal coordinates, spatial coordinates are represented in a two-dimensional structure using longitude and latitude, which are not easy to directly process. The original longitude and latitude coordinates $(x^{raw}_{lng}, y^{raw}_{lat})$ of each business district center are known, where $x^{raw}_{lng}$ represents the original longitude value of the business district and $y^{raw}_{lat}$ represents the original latitude value. First, we transform them into planar coordinates using the Mercator projection, denoted as $(x _{lng}, y_{lat})$. Then, we calculate the $M-1$ nearest neighbors for each business district using the Euclidean distance. Then, we scale the longitude and latitude for later position embedding. Therefore, the new coordinates are:
\begin{equation}
\begin{split}
    (x,y)^{scale}=(int(\frac{x_{lng}- min(\{x_{lng}\})}{max(\{x_{lng}\})-min(\{x_{lng}\})}\cdot N_x), \\
    int(\frac{y_{lat}- min(\{y_{lat}\})}{max(\{y_{lat}\})-min(\{y_{lat}\})}\cdot N_y)) \\
\end{split}
\end{equation}
where $\{\cdot\}$ denotes the set of all elements represented by $\cdot$ , for example, $\{x_{lng}\}$ represents the set of all values of $x_{lng}$ for business districts, and $int$ indicates rounding to the nearest integer. $N_x$ and $N_y$ represent the scaling factors for the coordinates $x$ and $y$, respectively. That means $x^{scale}\in [0,N_x)$ and $y^{scale}\in [0,N_y)$. Next, the spatial coordinates need to be transformed into relative coordinates, that is, the coordinates of surrounding business districts relative to the current business district:
\begin{equation}
(x_m,y_m)=(x^{scale}_{m}-x^{scale}_{center}+N_x, y^{scale}_{m}-y^{scale}_{center}+N_y)
\end{equation}
where $m\in [0,M)$ represents a specific business district. At this point, $x_m\in [0,2N_x)$ and $y_m\in[0,2N_y)$. Next, we can encode the spatial position like we do with temporal position, so:
\begin{equation}
SPE_{x_m}=OneHot(x_m)\cdot W_{SPE_x}
\end{equation}
\begin{equation}
SPE_{y_m}=OneHot(y_m)\cdot W_{SPE_y}
\end{equation}
where $W_{SPE_x} \in {R} ^{2N_x\times H}$ and $W_{SPE_y} \in {R} ^{2N_y\times H}$ are weight matrices for Spatial Position Embedding, randomly initialized. Finally, $SPE_{x_m}\in{R} ^H$ and $SPE_{y_m}\in {R} ^H$ are added together with spatial features.

\subsection{Spatial-Temporal Transformer} 
This module is designed primarily to encode spatio-temporal features and generate their summary. It mainly consists of the following two parts:

\textbf{Temporal Transformer}. This module encodes the temporal features of all business districts. For the temporal features of each business district, the Temporal Position Embedding is added to obtain the input of the Temporal Transformer:
\begin{equation}
\tilde{X} ^A_{m}=X^A_{m}W_{X^A}+TPE
\end{equation}
where $m\in [0, M)$ represents a specific business district, and $X^A_{m} \in{R} ^{N\times D_A}$ represents the A-type temporal features of business district $m$, and $W_{X^A}\in{R} ^{D_A\times H}$ represents its weight, mapping the input to the hidden size, enabling it to be added to $TPE\in{R} ^{N\times H}$. Then, $\tilde{X} ^A_m\in {R} ^{N\times H}$ goes through the Temporal Transformer, and the output is concatenated:
\begin{equation}
O^{temporal}=Concat(O^{temporal}_1,\cdots,O^{temporal}_m,\cdots ,O^{temporal}_M) \\
\end{equation}
\begin{equation}
O^{temporal}_m=transformer(\tilde{X} ^A_m)
\end{equation}
where $O^{temporal}\in {R}^{M\times H}$ is the output of the temporal transformer, and $H$ represents the hidden size of the transformer. $transformer(\cdot)$ utilizes the Transformer Encoder structure directly from the original paper \cite{vaswani2017attention}, and we skip the details of this process here.

\textbf{Spatial Transformer}.  This module encodes the spatial features abstracted by the Temporal Transformer. The output $O^{temporal}$ from the previous step is added to a 2D Spatial Position Embedding to obtain the input of the Spatial Transformer. After the Transformer processing, we obtain the output of the Spatial Transformer. The entire calculation process is as follows:
\begin{equation}
O^{spatial}=transformer(O^{temporal} + SPE_x + SPE_y)
\end{equation}

where $SPE_x\in{R}^{M\times H}$, $SPE_y\in {R}^{M\times H}$, and $O^{spatial}\in{R} ^H$. $H$ is the hidden size of the Spatial Transformer, which is kept the same as the Temporal Transformer. Here, the $transformer(\cdot)$ still uses the original transformer encoder structure without any modifications. Moreover, whether it is the Temporal Transformer or the Spatial Transformer, it is possible to configure multiple transformer layers, denoted as $L$.

\textbf{Spatio-Temporal Features}. In other words, for each time slice of each business district, there will be $D_{A}$ features. These features consist of four major categories: order volume, order acceptance rate, delivery completion rate, and delivery time, all of which are obtained by aggregating original order information to business districts. For each type of feature, we will calculate several different time granularities, such as: 5 minutes, 10 minutes, 30 minutes, 1 hour. The complete list of features can be found in Table~\ref{tab1}.

\begin{table}[htpb]
\caption{Detailed spatio-temporal features.}\label{tab1}
\begin{tabular}{l|p{1.3in}|p{2in}}
\toprule
Feature Type &  Subtype & Feature Name \\
\midrule
\multirow[b]{6}{*}{Number of order}  & Riders have not arrived at the store &  within the last hour, exceeding 0/8/15 minutes \\
& Riders have not yet picked up the meal &	within the last hour, exceeding 0/8/15 minutes \\
& Riders have not yet accepted the order & within the last hour, exceeding 0/8/15 minutes \\
& Number of uncompleted orders & in the last 10/30/60 minutes \\
& Number of completed orders & in the last 10/30/45/60 minutes \\
& Number of canceled orders & in the last 5/10 minutes \\
\hline
Order acceptance rate & Order acceptance rate & The acceptance rate in the last 3/5/10/15/30 minutes \\
\hline
\multirow[b]{2}{*}{Delivery completion rate} & Delivery completion rate & in the last 30/60 minutes \\
& On-time delivery completion rate & in the last 10/30/45/60 minutes \\
\hline
Delivery time & Average delivery time  & in the last 30/60 minutes \\
\bottomrule
\end{tabular}
\end{table}\textbf{}

The order acceptance rate refers to the number of orders not picked up by the riders divided by the total number of orders. During bad weather or peak periods, the riders may be overloaded and more likely to refuse or delay picking up orders. The delivery completion rate refers to the number of successfully delivered orders divided by the total number of orders. During peak periods, the user experience deteriorates, leading to an increase in order cancellations. It is important to note that the delivery time mentioned in the above features refers to the average delivery time of completed orders, which is different from the RPS label. The RPS label represents the delivery time for orders created in the next 5 minutes, which is currently unknown and needs to be predicted. These features all reflect the real-time status of the OFD logistics system, and may ultimately impact the delivery time of orders.
\subsection{Memory Network} 
\textbf{Memory Network}. This module is mainly used to encode peak periods, weather-related anomalies, as well as location, time, and other features. It incorporates shallow interactions based on attention mechanisms to enhance the model's sensitivity to these features. Our sensitive features are mainly categorical in nature. We handle them by passing them through an embedding layer separately, then concatenate the results and map them to obtain the input for the Memory Network:
\begin{equation}
\tilde{X} ^B=Concat(Emb_1(X^B_1), \cdots ,Emb_i(X^B_i), \cdots ,Emb_{D_B}(X^B_{D_B}))\cdot W_{X^B}
\end{equation}

where $i\in [0, D_B)$ represents the index of the i-th sensitive feature, $X^B_i$ represents the i-th feature, which is a categorical feature of type B (sensitive feature). $Emb_i$ represents the corresponding embedding function for the feature, which maps different $X^B_i$ to a fixed-dimensional embedding size $E$. $W_{X^B}\in {R} ^{D_BE\times H'}$ maps the output of the embedding of sensitive features to the hidden size of the memory network, denoted as $H'$.

For the memory network, the parameter matrix $W_{mem}\in {R} ^{L_{mem}\times D_{mem}}$ is a core parameter, where $L_{mem}$ is the number of stored patterns, set to 12, while $D_{mem}$ is the dimension of the patterns, set to 64. First, we concatenate $O^{spatial}$ and $\tilde{X} ^B$, and then calculate the query $q$:
\begin{equation}
q=[O_{spatial}\parallel \tilde{X}^B ]W_q+b_q
\end{equation}
where $\parallel$ represents concatenation, $W_q\in {R} ^{(H+H')\times D_{mem}}$ and $b_q$ represent the parameters of the query, therefore $q\in {R} ^{D_{mem}}$. Then, we calculate the attention based on $q$ and $W_{mem}$:
\begin{equation}
\alpha =Softmax(W_{mem}\cdot q)\times W_{mem}
\end{equation}
where $\alpha\in {R} ^{D_{mem}}$ represents the output of the attention. Finally, we concatenate $O^{spatial}$, $q$, and $\alpha$, input them into a multilayer perceptron (MLP), and obtain the final RPS estimation:
\begin{equation}
\hat{Y}=MLP(O^{spatial}\parallel q\parallel\alpha )
\end{equation}

\textbf{Sensitive Features}. This mainly includes the features of abnormal events, and it is important to ensure that RPS is sensitive enough to them. In addition to the peak periods and bad weather features mentioned earlier, we have also added city names and business district names, as the differences are often quite significant in different locations. Minute and the day of week have also been included, with the expectation that the model can learn differences at the minute level and fluctuations at the weekly level. The sensitive feature list is shown in Table~\ref{tab2}.

\begin{table}[t]
\caption{Detailed sensitive features.}\label{tab2}
\begin{tabular}{lp{3.7in}}
\toprule
Feature Name &  Examples \\
\midrule
City & Shanghai, Beijing, Hangzhou, etc. \\
Business district & People's Square business district, Xintiandi business district, etc. \\
Minute & 1-1440 \\
Peak period & Breakfast, Morning rush, Afternoon tea, Evening rush, Night snack \\
Day of week & 1-7 \\
Bad weather level & Normal weather, Slightly bad weather, Bad weather, Extremely bad weather \\
\bottomrule
\end{tabular}
\end{table}
The above features have all been converted into ID form.

\subsection{Model Training} 
We use the MAE loss function for end-to-end training of the entire model:
\begin{equation}
MAE=\frac{1}{N}\sum_{i=1}^{N}  \left | Y_i-\hat{Y}_i  \right |
\end{equation}
where $Y$ is the actual value of RPS corresponding to feature $X$, which is the average delivery time for orders placed in the business district in the next 5 minutes, as mentioned in section 3.

\section{Experiments}
\subsection{Experimental Setup}

\textbf{Dataset}. The dataset of Table~\ref{tab3} comes from Ele.me, including one month of data from Shanghai from October 5, 2023 to November 5, 2023, with 738,047 records, involving approximately 280 business districts. The dataset is divided into 3 parts: the training set contains 643,711 samples, involving 286 business districts, with a time range from October 5th to November 1st; the validation set includes 23,341 samples from November 2nd, involving 272 business districts; the test set contains 70,995 samples, involving 277 business districts, with a time range from November 3rd to November 5th. There are a total of 286 business districts in Shanghai, the training set includes all business districts, and the validation set and test set do not have records for some business districts because no orders were placed in these districts from November 2nd to November 5th. For the missing business districts, they may serve as the nearest neighbors for other business districts. We use a vector filled with -1 as the feature for these missing business districts.

\begin{table}[htbp]
\caption{Dataset Summary.}\label{tab3}
\centering
\begin{tabular}{lccc}
\toprule
 &  Training &  Validation & Testing\\
\midrule
\# of samples & 643711 & 23341 & 70995\\
\# of business district & 286 & 272 & 277 \\
\# of days & 5, Oct. - 1, Nov. & 2, Nov. & 3, Nov. - 5, Nov. \\
\bottomrule
\end{tabular}
\end{table}

\textbf{Baselines}. We select the following methods as baselines, which are widely used in the industry: Extreme Gradient Boosting (XGB) \cite{chen2015xgboost} was widely deployed online in the early days; DeepFM \cite{guo2017deepfm} is a classic deep learning based method, but it cannot use historical and spatial information, i.e., it only uses instantaneous features of the current business district; Long Short-Term Memory (LSTM) \cite{hochreiter1997long} and Graph Attention Networks (GAT)  \cite{velickovic2017graph} are commonly used sequential or graph models, respectively, which are good at handling time series or spatial graph structures, but they cannot handle spatio-temporal data at the same time.

\textbf{Metrics}. The most widely used metrics for evaluating regression performance include: Mean Absolute Error (MAE) and Mean Square Error (MSE). MAE measures the average absolute error between predicted and true values, while MSE is more sensitive to outliers and can to some extent measure the sensitivity of our model. In fact, we are more interested in Anomaly Mean Absolute Error (AMAE), calculated as follows:
\begin{equation}
AMAE=\frac{1}{N^\ast }\sum_{i=1}^{N^\ast }  \left | Y_i-\hat{Y}_i  \right | 
\end{equation}
where $i\in N^*$ represents the samples for which $Y_i \ge 40$ minutes, and $N^*$ represents the number of such samples. During events such as severe weather or peak periods, the RPS can exceed 40 minutes.

\textbf{Model setting}. Table~\ref{tab4} contains the core parameters and settings. In addition, we set the number of layers for the transformer to 1, dropout rate to 0.1. The model uses the Adam optimizer, with a learning rate of 0.001, and iterates over 1 epoch with a batch size of 512.

\begin{table}[htpb]
\caption{Core Parameters of the Model.}\label{tab4}
\begin{tabular}{l|c|c|p{2.3in}}
\toprule
Module & Parameter &  Value  & Definition \\
\midrule
\multirow[b]{5}{*}{Spatial-Temporal Features}  & $D_A$ &  31 & The number of spatial-temporal features (Type A features) is shown in Table 1.  \\
& $N$ & 6 & $N$time slices in the time dimension, sampled every 10 minutes. \\
& $M$ & 10 & $M$ business districts in the spatial dimension, composed of the current business district and its $M-1$ nearest neighbors. \\
& $N_x$ & 10 & The scaling factor for the spatial coordinate $x$. 
\\
& $N_y$ &10 & The scaling factor for the spatial coordinate $y$.\\
\hline
Sensitive Features & $D_B$ & 6 & The number of sensitive features (Type B features) is shown in Table 2. \\
\hline
\multirow[b]{2}{*}{Transformer} & $H$	& 256 & Transformer	Hidden size of the spatial transformer and temporal transformer. \\
& $L$ &	1& Number of layers in the transformer. \\
\hline
\multirow[b]{4}{*}{Memory Network}  &$E$& 8 &The embedding size of each category feature in sensitive features. \\
& $H'$	& 256 &	The hidden size of the memory network. \\
& $L_{mem}$	& 12 & Number of patterns stored in memory network. \\
& $D_{mem}$	& 64 & The dimension of the patterns in the memory network. \\
\bottomrule
\end{tabular}
\end{table}\textbf{}

\subsection{Experimental Results}
Table 5 shows the evaluation results of different methods. It can be seen from the table that the XGB method performs the worst, while our method STTM demonstrates the best performance. In general, deep learning based methods are generally superior to traditional methods, demonstrating the superiority of deep learning based methods in capturing feature dependencies. Furthermore, among the deep learning based methods, LSTM and GAT outperform DeepFM, as they either utilize temporal features or spatial features. 

In summary, compared to all the methods mentioned above, our STTM model achieved the best performance on all metrics, surpassing the best competitor by 9.66\%, 14.13\%, and 7.41\% in MAE, MSE, and AMAE, respectively. This is mainly because STTM not only leverages our specially designed powerful transformer architecture to encode spatio-temporal features, but also maintains sensitivity to abnormal events through a memory network.

\begin{table}[htpb]
\caption{Evaluation results of different models.}\label{tab5}
\centering
\begin{tabular}{l|c|c|c|c}
\toprule
Method & Model &  MAE  & MSE & AMAE \\
\midrule
Conventional & XGB & 2.9802&15.7228 & 9.2698 \\
\hline
\multirow{4}{*}{Deep learning based}  & DeepFM & 2.9129 & 15.1789 & 11.0193 \\
& LSTM & 2.8064 & 15.1553 & 11.7253 \\
& GAT & 2.8285 & 14.2588 & 9.9062 \\
& STTM  & \textbf{2.5352}  & \textbf{12.2442} & \textbf{8.5826} \\
\bottomrule
\end{tabular}
\end{table}

\subsection{Ablation Study}
To validate the effectiveness of each module in our design, we created some variants of the model. Specifically, we removed the Temporal Position Embedding, denoted as "w/o Temporal Position Embedding". Similarly, "w/o Spatial Position Embedding" means removing Spatial Position Embedding, "w/o Temporal Transformer" means removing Temporal Transformer, "w/o Spatial Transformer" means removing Spatial Transformer, and "w/o Memory Network" means removing Memory Network.

\begin{table}[htpb]
\caption{Ablation study results.}\label{tab6}
\centering
\begin{tabular}{l|c|c}
\toprule
 Model &  MAE  & MSE \\
\midrule
w/o Temporal Position Embedding & 2.7008 & 12.6327 \\
w/o Spatial Position Embedding & 3.2204 & 18.8824 \\
w/o Temporal Transformer & 2.6362 & 12.3175 \\
w/o Spatial Transformer  & 5.511 & 42.7432 \\
w/o Memory Network  & 2.6988 & 13.6346 \\
\hline
STTM  & \textbf{2.5352}  & \textbf{12.2442} \\
\bottomrule
\end{tabular}
\end{table}

As shown in Table~\ref{tab6}, STTM outperforms its best competitor by 3.83\% in MAE and 0.6\% in MSE. This means that when any component is missing, the model's performance deteriorates. This validates the effectiveness of each module in STTM, demonstrating that Temporal Position Embedding, Spatial Position Embedding, Temporal Transformer, Spatial Transformer, and Memory Network all contribute to the prediction.

\subsection{Parameter Analysis}
\begin{figure}
    \centering
    \includegraphics[width=1\linewidth]{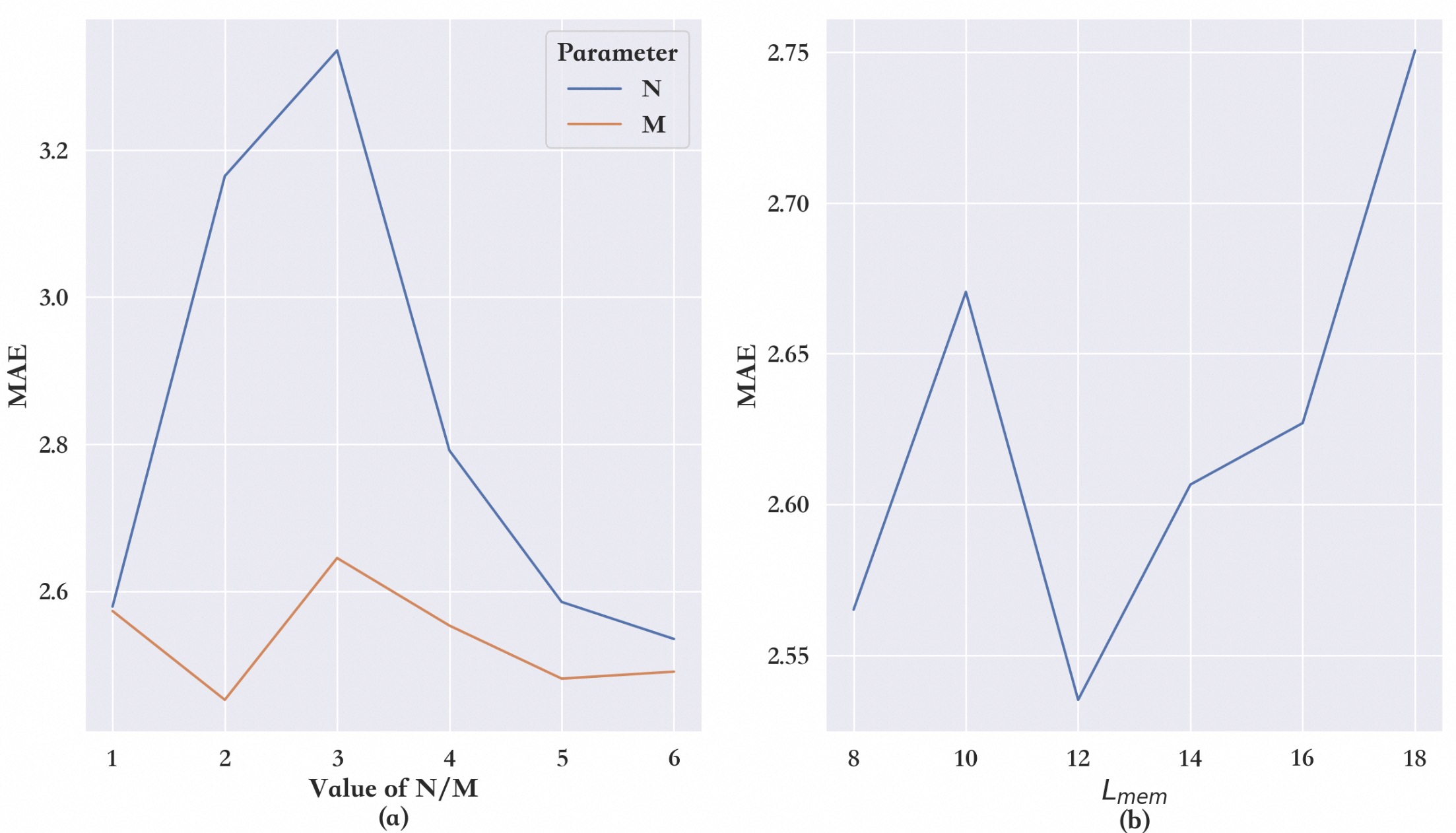}
    \caption{The impact of hyper-parameters on the performance of the model. (a) is the result of hyper-parameters N and M, and (b) is the result of $L_{mem}$.}
    \label{fig3}
\end{figure}

In this section, we explore the impact of the hyper-parameters $N$ (number of time slices in the temporal dimension), $M$ (number of business districts in the spatial dimension), and $L_{mem}$ (number of patterns in the Memory Network) on the model. During the experiment, only one parameter is changed at a time, while the other parameters remain constant. The results are shown in Fig.\@ \ref{fig3}.

First, we tested the impact of $N$ and $M$ by setting their values from 1 to 6, with an interval of 1. Basically, the performance of the model will increase first with the increase of $N$ and $M$, and then decrease. This indicates that adding more and more spatio-temporal information may initially be considered as noise, and later it becomes more useful for the model. In addition, we also modified $L_{mem}$ from 8 to 18, with an interval of 2. It can be observed that too many or too few patterns will lead to a decrease in the model's performance. The model performs best when $L_{mem}=12$.

\section{Conclusion}
In this paper, we propose a model based on Spatio-Temporal Transformer and Memory Network to predict real-time pressure signal, which effectively captures the unique spatio-temporal structures in the OFD domain and enhances its sensitivity to anomalous scenarios. The real-world logistics dataset experiments have demonstrated the effectiveness of our model, and this method has been successfully deployed in the online system of Ele.me.

\bibliographystyle{splncs04}
\bibliography{reference}

\end{document}